\documentclass[runningheads]{llncs}
\usepackage{svg}
\usepackage{graphicx}
\usepackage[export]{adjustbox}
\usepackage{enumerate}
\usepackage{amssymb}
\usepackage{cite}
\usepackage[subrefformat=parens,labelformat=parens]{subfig}
\usepackage{amsmath}
\usepackage{multirow}
\usepackage{booktabs}
\usepackage{textgreek}
\usepackage{hhline}
\usepackage{array}
\usepackage{tabu}
\usepackage{verbatim} 
\usepackage{graphics}
\usepackage{color}
\usepackage{url}
\usepackage[colorlinks = true]{hyperref}

\begin{document}
\title{A Few Shot Multi-Representation Approach for N-gram Spotting in Historical Manuscripts}

\titlerunning{Few Shot Multi-Representation N-gram Spotting for Historical Manuscripts}
 \author{
 Giuseppe De Gregorio\inst{1}\orcidID{0000-0002-8195-4118}\and
 Sanket Biswas\inst{2}\orcidID{0000-0001-6648-8270}\and
 Mohamed Ali Souibgui\inst{2}\orcidID{0000-0003-0100-9392}\and
 Asma Bensalah\inst{2}\orcidID{0000-0002-2405-9811}\and
 Josep Lladós\inst{2}\orcidID{0000-0002-4533-4739}\and
 Alicia Fornés\inst{2}\orcidID{0000-0002-9692-5336}\and
 Angelo Marcelli\inst{1}\orcidID{0000-0002-2019-2826}
 }

\authorrunning{G.De Gregorio et al.}
 \institute{
            DIEM - Department of Information and Electrical Engineering and Applied Mathematics \\ 
            University of Salerno, Italy\\
            \email{\{gdegregorio, amarcelli\}@cvc.uab.es} 
            \and
            Computer Vision Center \& Computer Science Department  \\
            Universitat Autònoma de Barcelona, Spain \\
 %          Fax: +123-45-678910\\
            \email{\{sbiswas, msouibgui, abensalah, josep, afornes\}@cvc.uab.es} 
}
% %

\maketitle              % typeset the header of the contribution
\begin{abstract}
Despite recent advances in automatic text recognition, the performance remains moderate when it comes to historical manuscripts. This is mainly because of the scarcity of available labelled data to train the data-hungry Handwritten Text Recognition (HTR) models. The Keyword Spotting System (KWS) provides a valid alternative to HTR due to the reduction in error rate, but it is usually limited to a closed reference vocabulary.
In this paper, we propose a few-shot learning paradigm for spotting sequences of a few characters (N-gram) that requires a small amount of labelled training data. We exhibit that recognition of important n-grams could reduce the system's dependency on vocabulary. In this case, an out-of-vocabulary (OOV) word in an input handwritten line image could be a sequence of n-grams that belong to the lexicon. An extensive experimental evaluation of our proposed multi-representation approach was carried out on a subset of Bentham's historical manuscript collections to obtain some really promising results in this direction.

\end{abstract}
\section{Introduction}\label{s:intro}

Historical document digitization is essential for preserving and maintaining the integrity of these documents' records. Handwriting recognition is a central task in facilitating different services related to historical manuscripts (e.g., searching, indexing, storing, etc.). In this context, handwritten historical documents processing is considered a challenging problem because images often suffer from degradation as a result of smears, artifacts, pen strokes, show-through, stains, bleed-through effects, and uneven illumination~\cite{antonacopoulos2007special}. A further obstacle is the writing style variability related to documents' writers living over different time periods~\cite{bunke2007off}. Moreover, most of these records have unique and complex structured layout patterns, which make them difficult to handle~\cite{lombardi2020deep}. This makes the usual recognition-based techniques for document processing (like OCR and HTR) not directly applicable to historical handwritten documents. The goal of an HTR system is to correctly map an input text image into a machine-encoded format at the word or character level. However, when writing is cursive, separation and recognition of characters is even more difficult~\cite{choudhary2013new}. Alternatively, the word-level prediction could be performed with greater accuracy in numerous cases. Hence, the KWS technique was introduced as a surrogate to recognition-based methods for word retrieval.

\begin{figure}[ht]
	\centering
    \includegraphics[width=0.25\columnwidth]{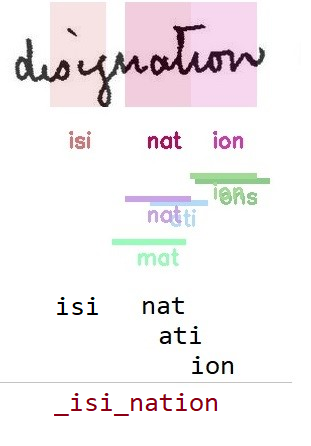}
	\caption{\textbf{Illustration of the few shot N-gram spotting task:} Given an OOV word text image as input, our model gives an almost correct recognition.}  
	\label{fig:example_spotting_01}
\end{figure}

KWS can be defined as the task of finding all the instances of a keyword in a document image without explicitly recognizing it \cite{stauffer2018keyword}. This approach makes the task more flexible and realizable, albeit the case when automatic transcription is impossible. Hence this technique is successful for documents of historical interest. The best performance of the systems mentioned above is achieved by lexicon-based systems~\cite{lombardi2020deep}. In contrast, the major drawback is their inability to correctly recognize words that are not lexicon elements~\cite{parziale2020one} (OOV words). This is a major constraint when building a scalable retrieval system, as obtaining labelled data is resource expensive and sometimes impossible in the case of historical documents.
The handwriting process can be considered a complex combination of motor and cognitive skills. As any intricate motor skill, it is acquired through the two learning principles: repetition and memorization. A neural scheme for motor learning has been proposed in~\cite{marcelli2013some}, where the authors state that sensory information is processed in the brain and appropriate motor commands are generated to execute the desired movement. During the learning phase of these movements, constant repetition enables the definition of motor automatisms that are subsequently activated when the same movement has to be performed again. Following the same approach to handwriting, we can presume that the handwriting learning process sets out an arrangement of automated movements corresponding to the best-learned writing motor primitives. The set of primitives depends on the writer's most familiar and best-learned sign sequence. 
This leads to the hypothesis that during the handwriting learning phase, an individual develops automated motor sequences to write the more frequent short sequences of characters, referred to as n-grams. Next, localizing these n-grams within a word could help with word-level recognition~\cite{poznanski2016cnn}. Thus, the ability to identify parts of a word could ease the recognition of the OOV words in a classical HTR system. 
Figure~\ref{fig:example_spotting_01} shows the effect of N-gram spotting on an OOV word. It can be seen how it is possible to partially recognise the word by focusing on the detection of some of its n-grams. These insights led us to develop an N-gram spotting system capable of recognizing partial sequences of words within a handwritten line of text.
On the other hand, current HTR models require large training sets. Nevertheless, finding historical documents together with manually labeled and aligned transcriptions is quite rare. Accordingly, a" few-shot" learning paradigm~\cite{wang2020generalizing} can be introduced and explored to train a recognition model. This kind of scenario allows the model to adapt to the limited training data provided. Following the ideas proposed in ~\cite{vinyals2016matching}, we can train a network to learn a similarity function between a handwritten text image query and a support N-gram image. Thus, this helps create a small training set composed of the most frequent n-grams extracted from available data collections. Later, we can detect n-grams inside handwritten documents of a test set in order to recognize text word images.

The main contributions of this work are summarized as follows:
\begin{itemize}
    \item We propose a few-shot N-gram spotting paradigm to deal with handwritten historical collections that have limited amount of transcription; 
    \item We explore and investigate the potential of the proposed N-gram spotting solution in KWS scenario of predicting OOV word samples in handwritten text images; 
    \item Correspondingly, a novel multi-representation fusion strategy of feature embedding attributes has been proposed for this task, to establish a tough-to-beat baseline.
\end{itemize}

The rest of the paper is organised as follows: Section~\ref{s:soa} explains the related work, Section~\ref{s:method} presents the architecture of the proposed method, while Section~\ref{s:results} reports the experimental results. Finally, Section~\ref{s:conclusion} presents the conclusions and future work.

\section{Related Work}\label{s:soa}
\noindent
\textbf{Keyword Spotting.} KWS was introduced to bridge the gap between digital and historical documents since conventional visual recognition systems can hardly process them.
In the classical approach, it is assumed that the text is already segmented into word candidates, and images of words are used as queries, referred to as Query-by-Example (QbE) in the literature. Thus the problem can be traced back to the definition of a distance measure between the different images of a single word~\cite{rath2007word}. Alternatively, segmentation-free systems do not provide any word-level segmentation, and the system searches within a line of text or on a whole page for regions similar to the example query~\cite{konidaris2016segmentation, biswas2021beyond, biswas2022docsegtr}. 
If the search query is fed to the system in the form of a string of characters and not as an image, it has been defined as Query-by-String (QbS). In this scenario, the system must be able to compare strings with images of candidate regions. An interesting solution for this task is to use an embedding to project both the query string and the document image features into a shared space where it is possible to define a distance function. In this common space, it is, therefore, possible to execute queries starting from both a string and an image~\cite{almazan2014word}. A powerful embedding function should be able to represent a word with its Pyramidal Histogram Of Characters (PHOC) attributes~\cite{almazan2013handwritten}. Using the PHOC representation, we could represent a generic string or a generic image word as a fixed-length vector by calculating the histogram of the characters of the input, concatenating the histogram of the first half and the second half, and so on until a certain depth is reached. Once the embedding space is defined, the search can be performed by learning a distance metric between the PHOC representations and then using the measure to spot keywords~\cite{sudholt2016phocnet}.
One set of KWS techniques that is particularly relevant in the case of historical documents is the lexicon-based approaches~\cite{puigcerver2017querying}. These systems are based on a list of predefined keywords that can be searched; hence any keyword not belonging to that list gives a null score. This stimulates the need for a robust KWS system that could provide a solution in the case of OOV keywords. In historical document analysis, the available data for training the recognition models is often very limited. In KWS, expanding the system lexicon and building large lexicons require large data sets. One method to mitigate the problem of finding OOV words is to define a similarity metric between generic words. In that way, it is possible to approximate the retrieval values of OOV words \cite{puigcerver2017querying}. Another approach is to use language models at the n-gram level, where the obtained predictions from the model are entrusted with the task of recognizing proposed words that are not in the lexicon \cite{kozielski2014open}. The performances of the proposed solutions do not exceed those of the lexicon-based systems, making the spotting of OOV words still an open problem.\\

\noindent
\textbf{Few Shot Learning for Handwriting Recognition.} Few shot learning setting is emerging in machine learning as an efficient parading to supervise the process with a reduced number of samples. One of the first papers that attempted to address the KWS problem with a few-shot approach was proposed by Howe et al. \cite{howe2013part}. Authors claimed their KWS system can perform the training phase with only a few labelled data. Since then, few-shot learning has shown a growing interest, and new solutions have been presented for handwriting analysis. Solutions ranged from generative systems capable of producing characters \cite{wong2015one} to character recognition systems in different languages \cite{shaffi2021few, wang2019radical}. Also, synthetic generation has been further expanded to generation of text-line images~\cite{kang2021content} and layouts in whole-page documents~\cite{biswas2021docsynth}.
One interesting solution was proposed by Souibgui et al. \cite{souibgui2021few}. The authors proposed a few-shot approach to search for a set of symbols within a handwritten line of an encrypted document. All above mentioned and other works demonstrate that few-shot can be fruitfully applied for handwriting analysis. 

\section{Methodology}\label{s:method}
Inspired by the work in~\cite{souibgui2021few}, we propose a few-shot learning-based model to tackle the N-gram spotting task in historical document collections. Given a handwritten input text image as a query and some of the most frequent N-gram examples as support, the model is trained to predict the position of the support N-grams within the query line image.

\subsection{The Base Architecture} \label{sect:3.1}

The proposed architecture is based on a faster R-CNN detector \cite{ren2015faster}. Then, it is adapted to a Siamese architecture by setting the goal of obtaining a similarity score between the images of the supporting N-grams and the boxes found within the line of handwritten text.
\begin{figure}[ht]
	\centering
	\includegraphics[width=0.9\columnwidth]{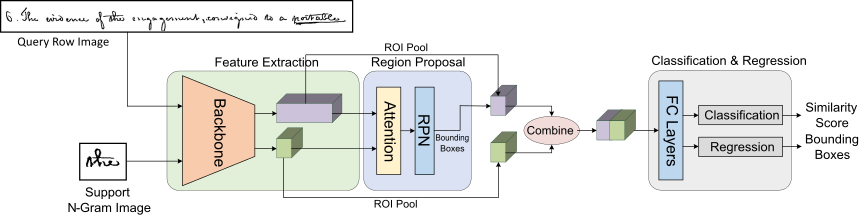}
	\caption{\textbf{Base Architecture.} A single backbone is used in feature extraction of the query and support images. The backbones for the query image and the support image share weights, as predicted by the Siamese model.}
	\label{fig:arch_01}
\end{figure}
Figure~\ref{fig:arch_01} shows the base architecture of the proposed system. Here, the supports are images of n-grams belonging to a given class $c$, and we form the model to detect all objects of class $c$ in a query image of a handwritten line of text $Q$. The first stage is a feature extraction phase where the CNN network extracts features from the text images. It must be highlighted that the network used to extract the features of both the query and the supporting images share the same structure and the same weights, as required for a Siamese architecture. the choice of a feature extraction backbone is not fixed, and by changing the choice of the feature extraction network, a different search modality can be applied which allows performing the spotting in a search space each time different.

The region proposal phase follows the feature extraction stage. In this phase, the feature map of both the query image and the supporting n-gram image are fed to an attention module and, further, to a region proposal network (RPN). %An attention mechanism upstream of the RPN network strengthens the region proposal phase by exploring the connection between the supporting image and the query image~\cite{fan2020few}. 
After this phase, ROI-pooling is introduced on top of the regions proposed by the RPN and the feature map of the supporting n-gram image.
The feature maps are then combined and then fed to a classification and regression module. This module consists of a collection of fully connected layers divided into two heads. 
The first is a classification head, and the second is a regression head. The output of the classifier uses a sigmoid activation function. The sigmoid decides whether the proposed region belongs to the class of the supporting image (1) or not (0). In parallel, the regression model generates the coordinates of the bounding boxes within the handwritten line image with respect to the classified image parts.

\subsection{The Multi-Modal Architecture} \label{sect:3.2}
\begin{figure}[ht]
	\centering
	\includegraphics[width=0.9\columnwidth]{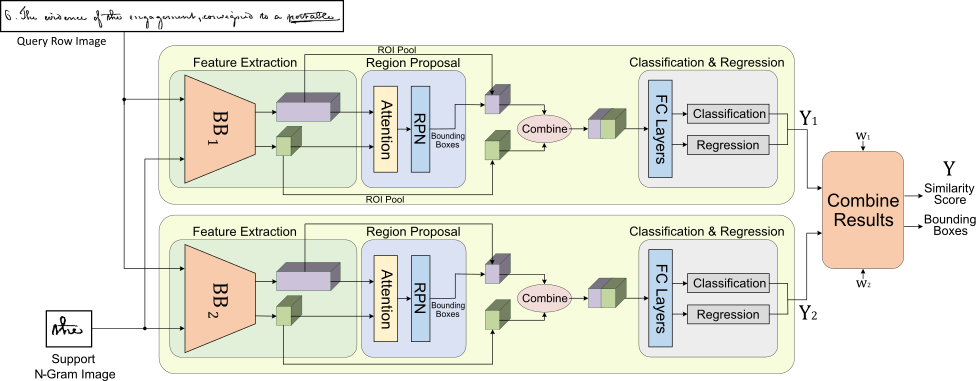}
	\caption{\textbf{The Multi-Modal Architecture:} Two independent solutions from different search modalities are fused.}
	\label{fig:arch_02}
\end{figure}

The base architecture can be constructed in by using different types of backbones for the feature extraction phase, so that different search modalities can be combined with the aim to fuse the results obtained in different feature spaces to obtain an improved spotting performance. 
In this section, we propose an architecture that combines two different search modalities as shown in Figure~\ref{fig:arch_02}, to which we refer to from now on as \emph{multi-modal architecture}. The two independent branches work concurrently, obtaining the two solutions $Y_1$ and $Y_2$ by using the backbones $BB_1$ and $BB_2$ respectively. At the end the two solutions are combined by a weighted concatenation as shown in eq.~\ref{eq:1conc}. 

\begin{equation}
Y = (w_{1} \cdot Y_1) || (w_{2} \cdot Y_2 )
\label{eq:1conc}
\end{equation}

Through the selected weights of each backbone $w_{1}$ and $w_{2}$, it is possible to define the relative importance of solution $Y_1$ in comparison to $Y_2$.

\begin{figure}[ht]
	\centering
	\includegraphics[width=0.3\columnwidth, height=3cm]{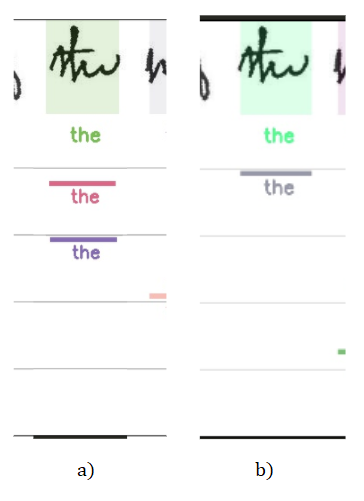}
	\caption{\textbf{An example of the combination of two similar solutions proposed for the same text area:} a) shows two proposal of the 3-gram "the" with two different scores; b) shows the result of the fusion. It results with a new interpretation for the 3-gram "the" whose score is higher than that of the initial interpretations.}
	\label{fig:fuse_exp}
\end{figure}

In the final combined solution $Y$, a possibility would be that for the same region of the handwritten query line image, the multi-modal system could propose multiple interpretations of the same N-gram as illustrated in Figure~\ref{fig:fuse_exp}. In this case, the similar solutions are fused into a new unique solution whose score is recomputed by first computing the gain and then adding it to obtain the final score. When two overlapping solutions $\alpha_1$ and $\alpha_2$ belonging to the same N-gram class are detected, they are fused together, computing a new score $s$ as in eq.~\ref{eq:3score} defined by the maximum between the two scores $s_1$ and $s_2$ incremented by a gain $\gamma$. 
\begin{equation}
s = max(s_1, s_2) + \gamma
\label{eq:3score}
\end{equation}
where the additional gain $\gamma$ is defined as:
\begin{equation}
\gamma = \delta \cdot \left(1 - \frac{|s_1 - s_2|}{max(s_1,s_2)}\right)
\label{eq:2score}
\end{equation}
Here in eq.~\ref{eq:2score}, $\delta$ is the maximum increment step, and $s_1$ and $s_2$ are the scores computed in the two solutions $\alpha_1$ and $\alpha_2$. Thus, the new score $s$ is directly proportional to the maximum score between $s_1$ and $s_2$ and to the difference between the two scores. Consequently, the cases where both branches propose the same interpretation with very high scores for the same text region are rewarded. If the new score exceeds the maximum allowable score value (in our case $s>1$), the score of the recomputed solution is set equal to one, while the scores of all other interpretations that overlap for the same range of the query image are decreased by the excess value $1-{s}$.

\subsection{Multi-Modal Architecture with Early Fusion} \label{sect:3.3} 
\begin{figure}[ht]
	\centering
	\includegraphics[width=0.9\columnwidth]{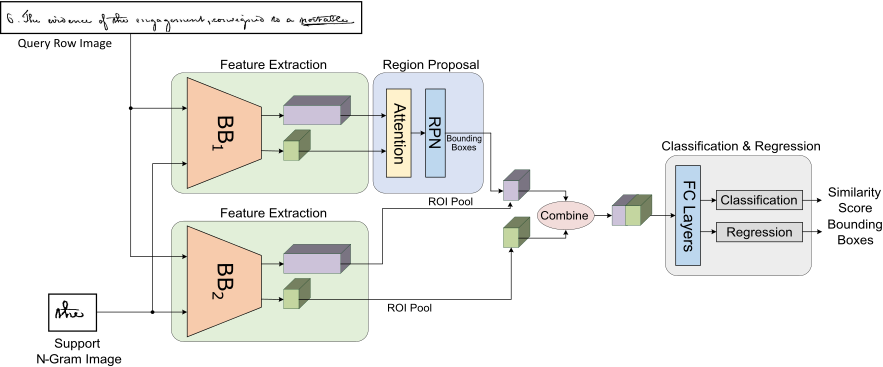}
	\caption{\textbf{The Multi-Modal Architecture with Early Fusion:} The architecture uses different modalities for region proposal and classification and regression.}
	\label{fig:arch_03}
\end{figure}

The previous multi-modal architecture performs the same region proposal, classification, and regression problems twice in both branches, using two different backbones before fusing the results at the end to obtain the final scores. Instead, an early fusion strategy that uses one branch for region proposal and the other branch for classification and regression could also be adopted. Thus, we present a modification of the previous multi-modal architecture as shown in Figure.~\ref{fig:arch_03}. The key objective is to separate the feature space for the region proposal problem from the feature space for the classification and regression problem. Therefore, the first branch is used for the region proposal phase. It computes the features of the query line images and supporting n-grams used by the region proposal block. Then, it identifies the possible regions of the text row that are candidates for an instance of the supporting image N-gram. Afterward, the second branch calculates the features used in the classification and regression phase, using the regions proposed in the other branch.
 
\section{Experiments}\label{s:results}
\subsection{Experimental Setup}

Among the premises of this work, the most restrictive is related to the amount of available training data. Indeed, our application scenario consists of very small handwritten document collections. Thereby, we considered a subset of the well-known Bentham Collection used in~\cite{sanchez2015icdar}. 20 pages were selected, 5 of which were used as a training set. The entire transcription and examples of frequent reference n-grams were extracted from this training set, obtaining a list of key n-grams consisting of 116 classes, each populated by a number of items ranging from five to seventy-six. The remaining 15 pages were used as a test set. All pages of the dataset were pre-processed by concatenating a binarization step \cite{sauvola2000adaptive},  a text line segmentation, and lastly, a deslanting method~\cite{vinciarelli2001new}.

Since limited training data is available, it may be useful to use different data sets for pre-training. In our case, we have augmented the pre-training dataset with two additional datasets. Additionally, we have created two synthetically generated datasets that are much larger than our real data.
To create the first synthetic dataset, we have chosen the Omniglot dataset \cite{lake2015human}. Omniglot consists of 1623 different characters handwritten by different scribes from 50 different alphabets. There are 20 examples of each character in this dataset. We generated 2000 lines of text with 964 different symbols by randomly spacing the symbols in the series with a high probability of symbol overlap. Next, for the second set of synthetic training data, lines of handwritten text were instead generated using a generative network capable of generating words from handwritten text~\cite{kang2020ganwriting, kang2020distilling}. The network was trained on the IAM dataset \cite{marti2002iam} and then used to generate random words (assembled into 2000 lines of text). In this case, it was possible to use these lines to highlight the contained n-grams and to use these n-grams as marked elements for the training set.

\subsection{Evaluation Metrics}

\begin{figure}[ht]
	\centering
	\includegraphics[width=0.9\columnwidth]{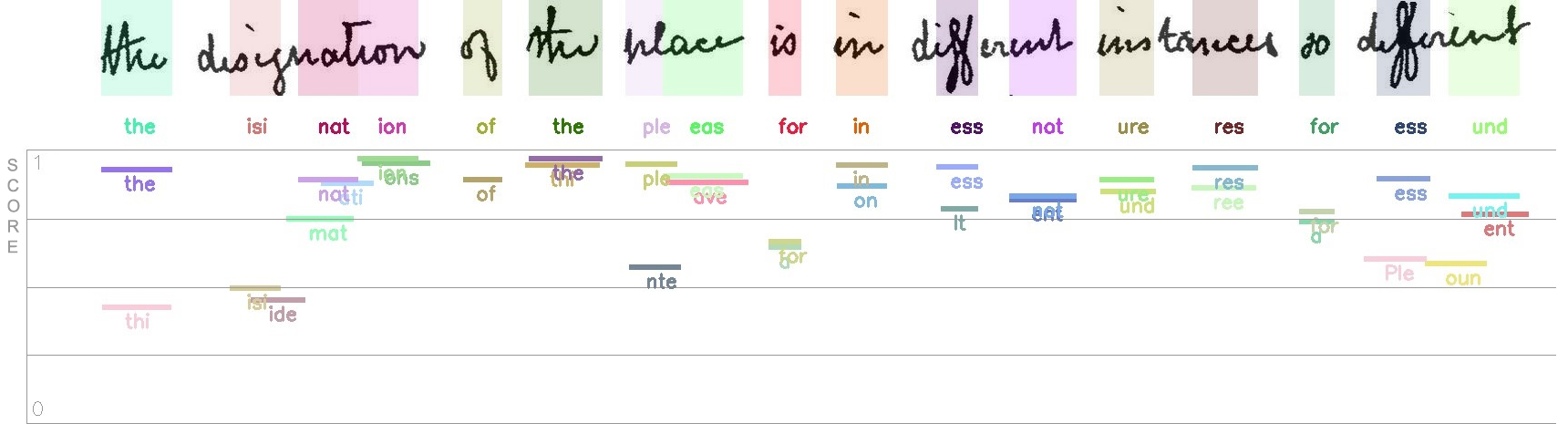}
	\caption{\textbf{Example of the result of the spotting on a query text line image:} the image highlights all the boxes detected in correspondence with each n-gram reporting the scores associated with each box at the bottom.}
	\label{fig:res_01}
\end{figure}

As observed in the Figure ~\ref{fig:res_01}, the system offers different interpretation options for each text area of the text line, each with a different similarity score value between 0 and 1. However, the system obtains a single interpretation for each class of n-grams for each text line area. This is due to the options merging of the multi-modal architecture module described in Section \ref{sect:3.2}. We address the task as a retrieval problem. Thus, the system is formulated as a multiple item recommendation system, with the peculiarity that the system cannot propose multiple options belonging to the same class. On the other hand, our interest lies in the fact that at least one of the top-k options proposed by the system is the correct one; thus, the transcription of the N-gram has been correctly identified. We then propose a modification of the Precision and Recall at $k$ metrics as shown in eq.~\ref{eq:4metric} and eq.~\ref{eq:5metric} respectively.\\
\begin{equation}
p@k = \frac{true\_relevant\_ngrams@k}{retrieved\_ngrams}
\label{eq:4metric}
\end{equation}
\begin{equation}
r@k = \frac{true\_relevant\_ngrams@k}{relevant\_ngrams}
\label{eq:5metric}
\end{equation}
In the aforementioned formulations, $true\_relevant\_ngrams@k$ represents the number of N-grams correctly detected given the top $k$ options for each area of the text line image, $retrieved\_ngrams$ denotes the number of all n-grams detected within the text line, while $relevant\_ngrams$ is the number of n-grams that make up the text line.
It should be noted, however, that the system is not able to recognise n-grams outside the list of n-grams used for the search in the test phase but that it proposes an interpretation for all areas of the queried text line in any case, even if it has a low score. Therefore, it is interesting to limit the analysis to the n-grams that the system can actually recognise. We will refer to these n-grams by the term \emph{in vocabulary}. In addition, we use a recall metric at $k$ restricted to n-grams in the vocabulary, defined as in eq.~\ref{eq:6metric} where the $relevant\_ngram\_InVoc$ term denotes the number of n-grams in the vocabulary that occur in the handwritten text line image.
\begin{equation}
r@k\_InVoc = \frac{true\_relevant\_ngrams@k}{relevant\_ngrams\_InVoc}
\label{eq:6metric}
\end{equation}

\subsection{Results and Discussion}

\begin{figure}[t]
	\centering
	\begin{tabular}{ccc}
	    \subfloat[]{\includegraphics[width = 1.25in]{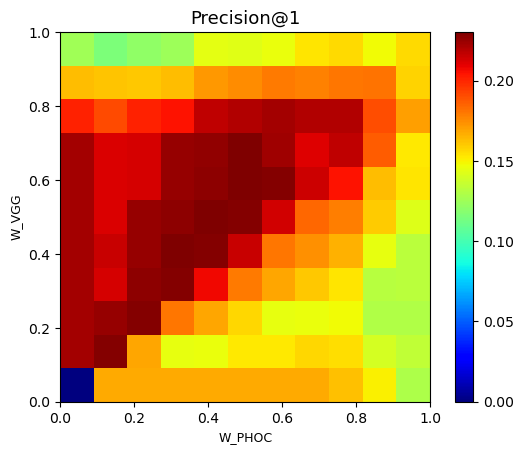}} &
	    \subfloat[]{\includegraphics[width = 1.25in]{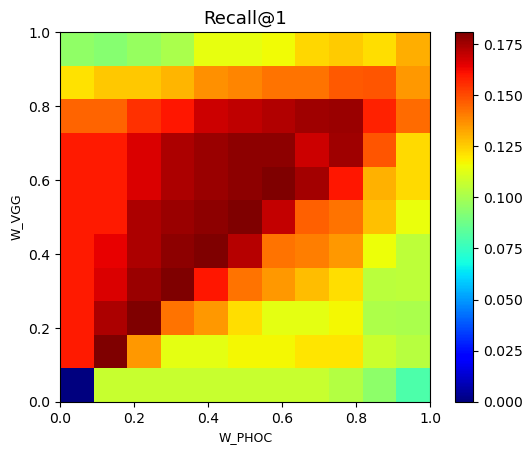}} &
	    \subfloat[]{\includegraphics[width = 1.25in]{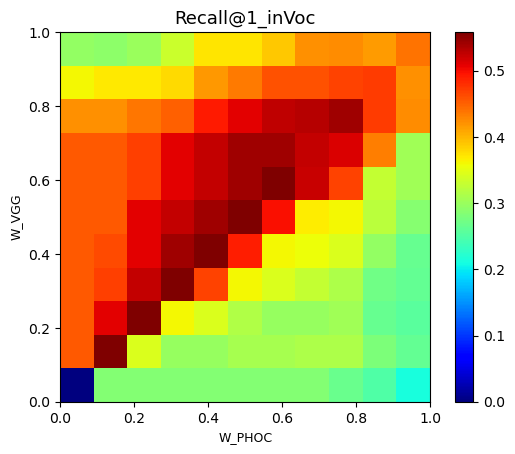}} \\
	    
	    \subfloat[]{\includegraphics[width = 1.25in]{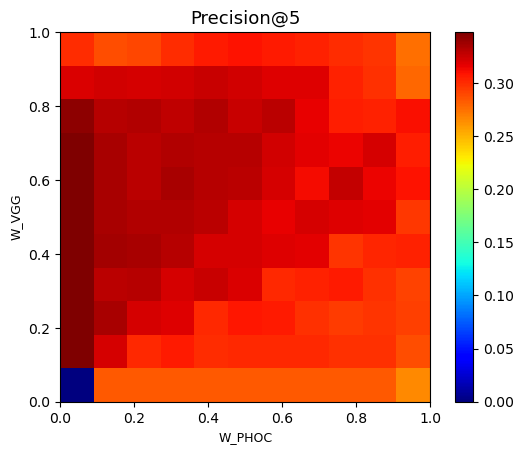}} &
	    \subfloat[]{\includegraphics[width = 1.25in]{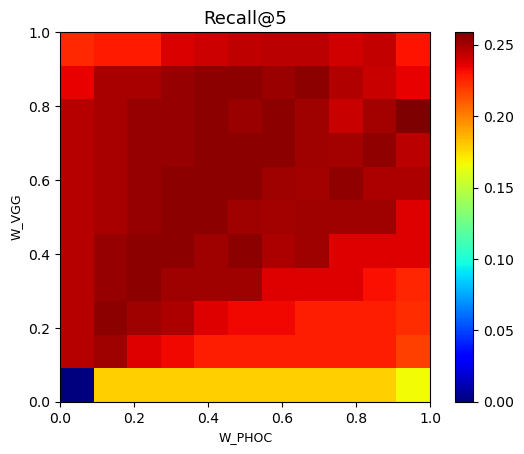}} &
	    \subfloat[]{\includegraphics[width = 1.25in]{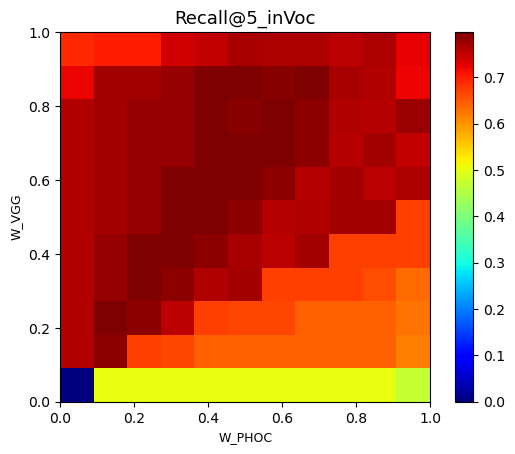}}

	\end{tabular}
	\caption{\textbf{Visualizing feature weights for Multi-modal Fusion.} Variation of the (a)Precision@1 (b)Recall@1 (c)Recall@1\_InVoc (d)Precision@5 (e)Recall@5 (f)Recall@5\_InVoc according to the fusion weights for visual and phoc features in case of 5-shot scenario.}
	\label{fig:rec_at_5_inv}
\end{figure}

\noindent
\textbf{Selecting Base Architecture Backbones:}
We selected several model backbones to test the base architecture and evaluated their performance. We have chosen to use two backbones related to two architectures that have performed well on different computer vision tasks: VGG16~\cite{simonyan2014very} and Resnet18~\cite{he2016deep}. We have also decided to test the PHOCnet~\cite{sudholt2016phocnet} backbone as it is widely used in various handwritten word spotting tasks. We compare our approach with the solution presented in ~\cite{souibgui2021few}. This work was designed to search for encrypted symbols in a line of handwritten text using a few-shot approach. Although there is a difference in domains, the commonality between these two problems is that both approaches have solutions based on a few-shot setting, and both solutions search for a symbol within a line of handwritten text image (whether it is an encrypted symbol or a Latin N-gram).
Table~\ref{tab:res_01} shows the results for a single branch architecture with different backbones compared to the results obtained with the model presented in~\cite{souibgui2021few}. The results are given for the cases $k=1$ and $k=5$ and for the number of shots of $1$, $3$, and $5$. The different systems have been trained with the set of n-grams extracted from the Bentham collection training data, while for the model \cite{souibgui2021few}, the training conditions that showed better performance were chosen. We can infer from the table how the model performance changes depending on the backbone network used for the feature extraction phase. As outlined, the VGG16 network backbone in our base model achieves the best results.\\
\begin{table}[ht]
\caption{Results for the base architecture evaluating different backbones ($BB$) for feature extraction:}
\label{tab:res_01}
\centering
\begin{tabular}{@{}ccccccccc@{}}
\toprule
\textbf{} & \textbf{\#shot} & \textbf{p@1} & \textbf{r@1} & \textbf{r@1\_InVoc} & \textbf{p@5} & \textbf{r@5} & \textbf{r@5\_InVoc} \\ \midrule
\multirow{3}{*}{Souibgui et al. \cite{souibgui2021few}}
        & 1 & 0.0325 & 0.0279 & 0.0557 & 0.0607 & 0.0513 & 0.1218  \\
        & 3 & 0.0269 & 0.0256 & 0.0598 & 0.0757 & 0.0717 & 0.1864  \\
        & 5 & 0.0122 & 0.0121 & 0.0348 & 0.0671 & 0.0641 & 0.1976  \\ \midrule
\multirow{3}{*}{${BB}_{VGG16}$}
        & 1 & \textbf{0.0701} & 0.0373 & \textbf{0.0942} & 0.1747 & 0.0854 & 0.2114 \\
        & 3 & 0.0579 & \textbf{0.0339} & 0.0638 & 0.1947 & 0.1006 & 0.2302\\
        & 5 & 0.0465 & 0.0274 & 0.0513 & \textbf{0.1980} & \textbf{0.1062} & \textbf{0.2732} \\ \midrule
\multirow{3}{*}{${BB}_{RESNET18}$}
        & 1 & 0.0000 & 0.0000 & 0.0000 & 0.0533 & 0.0061 & 0.0150 \\
        & 3 & 0.0333 & 0.0042 & 0.0167 & 0.0333 & 0.0042 & 0.0167 \\
        & 5 & 0.0167 & 0.0026 & 0.0059 & 0.0333 & 0.0061 & 0.0150 \\ \midrule
\multirow{3}{*}{${BB}_{PHOCnet}$}
        & 1 & 0.0000 & 0.0000 & 0.0000 & 0.0167 & 0.0048 & 0.0083 \\
        & 3 & 0.0310 & 0.0071 & 0.0392 & 0.0671 & 0.0153 & 0.0566 \\
        & 5 & 0.0111 & 0.0048 & 0.0083 & 0.0954 & 0.0233 & 0.0780 \\ \bottomrule
\end{tabular}
\end{table}
\noindent
\textbf{The Significance of Pre-training:} So far, the performance of the model seems unsatisfactory in most cases. To achieve a better performance, a pre-training phase is essential to be conducted using synthetic data. For this experimentation, we tested two backbones: the VGG16 backbone because it showed better performance compared to the architecture with the Resnet18 backbone, while the PHOCnet backbone was chosen due to its relevance to the application domain. Table \ref{tab:res_02} highlights the results and illustrates how the pre-training phase on synthetic data, followed by a fine-tuning phase on the real data, allows the system to significantly improve its performance in all possible metrics. It is worth mentioning that it was impossible to use the synthetic training data generated from the Omniglot dataset with the PHOCnet network since a PHOC representation of a word composed of symbols rather than letters is meaningless.\\
\noindent
\textbf{Comparison to existing SOTA methods:} The model from Souibgui et al.~\cite{souibgui2021few} has  been pre-trained with a dataset generated from the Omniglot dataset. For a fair comparison with our proposed approach, we have performed a fine-tuning phase on the real data. As denoted in Table~\ref{tab:res_02}, the results obtained with the model from Souibgui et al.~\cite{souibgui2021few} are comparable to the results of our model with the VGG16 backbone and the synthetic Omniglot dataset for pre-training. The model in~\cite{souibgui2021few} uses a VGG16 backbone for feature extraction, which could explain the comparable results. 
It is also interesting to note that the proposed system performs better in the case of $k=5$ and when the number of shots is high.\\ 
%This suggests that the contribution of the to recalculate the score of overlapping solutions actually improves the performance of the system.
% This suggests that recalculating the score of overlapping solutions actually improves the system's performance.
\begin{table}[ht]
\caption{Comparative results with the Single-Branch pre-trained model, the Multi-modal model, and the Multi-modal model with Early Fusion:}
\label{tab:res_02}
\centering
\resizebox{\textwidth}{!}{%
\begin{tabular}{ccccccccc}
\hline
\multicolumn{2}{c}{} & \textbf{\#shot} & \textbf{p@1} & \textbf{r@1} & \textbf{r@1\_InVoc} & \textbf{p@5} & \textbf{r@5} & \textbf{r@5\_InVoc} \\ \hline

%Souibgui
\multicolumn{2}{c}{\multirow{3}{*}{Souibgui et al.~\cite{souibgui2021few} + finetuning}} 
                     & 1 & 0.1613 & 0.1121 & 0.3408 & 0.3007 & 0.2042 & 0.6465 \\
\multicolumn{2}{c}{} & 3 & 0.2088 & 0.1427 & 0.4391 & 0.3436 & 0.2301 & 0.6948 \\
\multicolumn{2}{c}{} & 5 & 0.2241 & 0.1602 & 0.4703 & 0.3342 & 0.2364 & 0.7314 \\ \midrule

% Single Branch finetined
 & \multirow{3}{*}{${BB}_{VGG16}$ + Omniglot} 
                        & 1 & 0.1823 & 0.1291 & 0.3376 & 0.3126 & 0.2171 & 0.6092 \\
\multicolumn{1}{l}{} &  & 3 & 0.2060 & 0.1405 & 0.4279 & 0.3462 & 0.2349 & 0.7170 \\
\multicolumn{1}{l}{} &  & 5 & 0.2216 & 0.1470 & 0.4436 & 0.3725 & 0.2450 & 0.7594  \\ \cline{2-9} 
\textit{} & \multirow{3}{*}{${BB}_{VGG16}$ + Synth} 
                       & 1 & 0.1555 & 0.0778 & 0.2298 & 0.2940 & 0.1511 & 0.4908 \\
\textit{Single-Branch} &  & 3 & 0.1823 & 0.1036 & 0.3357 & 0.3019 & 0.1644 & 0.5346 \\
\multicolumn{1}{l}{} &  & 5 & 0.1878 & 0.1117 & 0.3292 & 0.3336 & 0.1856 & 0.5832 \\ \cline{2-9} 
\multicolumn{1}{l}{} & \multirow{3}{*}{${BB}_{PHOCnet}$ + Synth} 
                        & 1 & 0.1084 & 0.0612 & 0.1669 & 0.2682 & 0.1543 & 0.4156 \\
\multicolumn{1}{l}{} &  & 3 & 0.1369 & 0.0831 & 0.2105 & 0.2734 & 0.1638 & 0.4781 \\
\multicolumn{1}{l}{} &  & 5 & 0.1333 & 0.0791 & 0.1976 & 0.2861 & 0.1700 & 0.4809 \\ \midrule

% Multi-modal
\multicolumn{1}{l}{} & \multirow{3}{*}{${BB}_{VGG16}+{BB}_{PHOCnet}$} 
                         & 1 & 0.2140  & 0.1424 & 0.4505 & 0.3437 & 0.2392 & 0.7195 \\
\textit{Multi-modal}  &  & 3 & 0.2045 & 0.1495 & 0.4783 & 0.3536 & 0.2519 & 0.7498 \\
                      &  & 5 & 0.2303  & \textbf{0.1808} & \textbf{0.5582} & 0.3747 & \textbf{0.2588} & \textbf{0.7975} \\ \midrule

% Multi-modal early-fusion
                          &  & 1 & 0.0596 & 0.0553 & 0.1563 & 0.1504 & 0.1470 & 0.4027 \\
 & VGG16 ROI + PHOCnet class & 3 & 0.0793 & 0.0832 & 0.2192 & 0.1552 & 0.1619 & 0.4199 \\
\textit{Early-Fusion}     &  & 5 & 0.0536 & 0.0582 & 0.1652 & 0.1715 & 0.1858 & 0.5105 \\ \cline{2-9} 
\textit{Multi-modal} &       & 1 & 0.2346 & 0.1179 & 0.3514 & 0.3374 & 0.1738 & 0.5762 \\
 & PHOCnet ROI + VGG16 class & 3 & 0.2234 & 0.1312 & 0.4258 & 0.3783 & 0.2185 & 0.6925 \\
                          &  & 5 & \textbf{0.2419} & 0.1392 & 0.4404 & \textbf{0.4088} & 0.2236 & 0.7043  \\ \midrule
\end{tabular}
}\end{table}
\noindent
\textbf{Effectiveness of Multi-Modal Architecture:} 
In this section we evaluate the model described in Section \ref{sect:3.2} using the two different backbones VGG16 and PHOCnet, respectively pre-trained with the dataset generated from the Omniglot dataset and with the synthetically generated dataset containing n-grams. 
The outcome of the model depends on the values assigned to the parameters $w_1$ and $w_2$ of the eq.~\ref{eq:2score}. Next, we evaluated the model by varying the two weights from the minimum value 0 to the maximum value 1 with a step size of 0.1 and testing all the weighs combination by performing a grid search ~\cite{bergstra2012random}. Fig.~\ref{fig:rec_at_5_inv} shows the results of the search, and it is clear that the best results are obtained when combining the two different branches. Table~\ref{tab:res_03} illustrates the best results for each metric in case of 5-shot scenario. As an overall trend, the weighting of the branch with VGG16 backbone is always greater than or equal to the weighting of the branch with PHOCnet backbone, except in the case of $r@5$. Based on the weights values, we can conclude that more discriminative feature spaces should have greater importance for combination purposes, complementing our intuition. Additionally, the performance of all indices is higher than the previous performance with the single backbone model. This suggests that searching over multiple feature spaces may contribute to better performance. 
\\
\begin{table}[!ht]
\caption{Best recorded results with relative weights.}
\label{tab:res_03}
\centering
\begin{tabular}{lccc}
\hline
\multicolumn{1}{c}{\textbf{Measure}} & \textbf{Max-Value} & \textbf{w\_vgg16} & \textbf{w\_phoc} \\ \hline
\textit{Precision@1} & 0.2303 & 0.7 & 0.5 \\
\textit{Recall@1} & 0.1808 & 0.5 & 0.5 \\
\textit{Recall@1\_InVoc} & 0.5582 & 0.4 & 0.4 \\
\textit{Precision@5} & 0.3747 & 0.5 & 0 \\
\textit{Recall@5} & 0.2588 & 0.8 & 1.0 \\
\textit{Recall@5\_InVoc} & 0.7975 & 0.6 & 0.3 \\ \hline
\end{tabular}
\end{table}
\noindent
\textbf{Early fusion strategy:} As discussed in Section \ref{sect:3.3}, it may be interesting to investigate the early fusion strategy during the multi-modal combination of the two backbones. To this end, we have trained the model of Section~\ref{sect:3.3} again using the two pretrained VGG16 and PHOCnet backbones, as in the case of the previous section. The architecture has been first implemented with the VGG16 backbone for the region proposal phase and the PHOCnet backbone for the classification phase, then with the PHOCnet backbone for the region proposal and the VGG16 for the classification. Table~\ref{tab:res_02} presents information on the obtained results. According to Table~\ref{tab:res_02}, the performance of the system decreases compared to the best shots with the previous solutions in terms of recall but increases in terms of precision. This entails that the two different strategies for the fusion are better suited to different scenarios, depending on whether the goal is to maximize recall rather than precision.

\section{Conclusion}\label{s:conclusion}
In this paper, we have presented a few-shot multi-representation model for N-gram spotting for historical manuscripts.
Based on experimental results, we have demonstrated our approach's adequacy and that it is possible to detect n-grams of references within a line of handwritten text.
We limited the analysis to some of the available feature spaces, but the results showed that using more than one representation can improve performance. In this direction, an exhaustive study on the different representations available could be performed by analyzing in detail different backbones architecture with different depths to understand the impact on the model.
The method does not limit the search for n-grams only to the words that appear in the training pages, from which classical recognition systems build their reference lexicons, but offers the possibility to recognise n-grams also in the OOV words that appear only in the test pages. 
In the future, we will explore the possibility of extending the model with a "word reconstruction" module based on a language model that can reconstruct whole-word interpretations starting from the recognised n-grams to bring the focus of the problem back to the most useful word-level. 
We will also explore the possibility of extending the model to the page level in order to avoid the segmentation of handwritten text line images.

 \section*{Acknowledgment}
% Hidden due to anonymous submission.
This work has been partially supported by the Spanish projects RTI2018-095645-B-C21, PID2021-126808OB-I00 and FCT-19-15244, and the Catalan projects 2017-SGR-1783, the CERCA Program / Generalitat de Catalunya, PhD Scholarship from AGAUR (2021FIB-10010), and the DIEM Graduate Research Scholarship entitled "Strumenti di supporto alla trascrizione di documenti manoscritti di interesse storico-culturale"

\bibliographystyle{splncs04}
\bibliography{main}




% LNCS Homepage, \url{http://www.springer.com/lncs}. Last accessed 4
% \end{thebibliography}
\end{document}